\documentstyle{elsart1p}
\newtheorem{theorem}{Theorem}
\newtheorem{proposition}{Proposition}

\newtheorem{lemma}{Lemma}

\begin{document}

\begin{frontmatter}


\title{On empirical meaning of sets of algorithmically 
random and non-random sequences}


\author{Vladimir V. V'yugin\thanksref{label1}}
\thanks[label1]{This paper was presented in part at
2nd International Computer Science Symposium in Russia.
Ekaterinburg, Russia, September 3-7, 2007~\cite{Vyu2007}. This
research was partially supported by Russian foundation for
fundamental research: 06-01-00122-a.}

\address{Institute for Information Transmission Problems,
Russian Academy of Sciences,
Bol'shoi Karetnyi per. 19, Moscow GSP-4, 127994, Russia.
e-mail vyugin@iitp.ru}

\begin{abstract}
We study the a priori semimeasure of sets of $P_\theta$-random 
infinite sequences, where $P_\theta$ is a family of probability distributions
depending on a real parameter $\theta$.
In the case when for a computable probability distribution
$P_\theta$ an effectively strictly consistent estimator exists,
we show that the Levin's a priory semimeasure of the set of all
$P_\theta$-random sequences is positive if and only if the
parameter $\theta$ is a computable real number.
For the Bernoulli family $B_\theta$,
we show that the a priory semimeasure of the set
$\cup_\theta I_\theta$, where $I_\theta$ is the set of all
$B_\theta$-random sequences and the union is taken over all
non-random $\theta$, is positive.
\end{abstract}

\end{frontmatter}

\maketitle

\section{Introduction}

We use algorithmic randomness theory to analyze ``the size'' of sets 
of infinite sequences random with respect to parametric families
of probability distributions. 

Let a parametric family of
probability distributions $P_\theta$, where $\theta$ is a real number,
be given such that an effectively strictly consistent estimator
exists for this family. The Bernoulli family with a real parameter
$\theta$ is an example of such family. Theorem~\ref{th-1} shows that
the Levin's a priory semimeasure of the set of all
$P_\theta$-random sequences is positive if and only if the
parameter value $\theta$ is a computable real number.

We say that a property of infinite sequences has no ``empirical meaning'' if
the Levin's a priory semimeasure of the set of all sequences possessing
this property is $0$. In this respect, the model of
the biased coin with ``a prespecified'' probability $\theta$ of head is
meaningless when $\theta$ is a noncomputable real number; noncomputable
parameters $\theta$ can have empirical meaning only in their totality, i.e.,
as elements of some uncountable sets. For example, $P_\theta$-random
sequences with noncomputable $\theta$ can be generated by a
Bayesian mixture of these $P_\theta$ using a computable prior.
In this case, evidently,
the semicomputable semimeasure of the set of all sequences random with
respect to this mixture is positive.

We give in Appendix~\ref{proof-1} the simple proof of our previous result
(formulated in Theorem~\ref{pr-1}) which says
that the Levin's a priory semimeasure of the set of
all infinite binary sequences non-equivalent by Turing to
Martin-L\"of random sequences is positive. In particular, 
these sequences are non-random with respect to each computable probability 
distribution. 

We use this result to prove Theorem~\ref{n-1}. This theorem 
shows that a probabilistic machine can be
constructed, which with probability close to $1$ outputs a random
$\theta$-Bernoulli sequence such that the parameter $\theta$ is not
random with respect to each computable probability distribution.
This result can be interpreted such that the Bayesian statistical approach
is insufficient to cover all possible ``meaningful'' cases for $\theta$-random
sequences.

\section{Preliminaries}\label{prel-1}

Let $\Xi$ be the set of all finite binary sequences, $\Lambda$ be
the empty sequence, and $\Omega$ be the set of all infinite binary sequences.
We write $x\subseteq y$ if a sequence $y$ is an extension
of a sequence $x$, $l(x)$ is the length of $x$. For any $\omega\in\Omega$,
$\omega^n=\omega_1\dots\omega_n$. A real-valued
function $P(x)$, where $x\in\Xi$, is called semimeasure if
\begin{eqnarray}
P(\Lambda)\le 1,
\nonumber
\\
P(x0)+P(x1)\le P(x)
\label{semi-1}
\end{eqnarray}
for all $x$, and the function $P$ is semicomputable from below;
this means that the set
$\{(r,x):r<P(x)\}$, where $r$ is a rational number, is recursively enumerable.
A definition of upper semicomputability is analogous.

Solomonoff proposed ideas for defining the a priori probability
distribution on the basis of the general theory of algorithms.
Levin~\cite{Lev73,ZvL70} gave a precise form of
Solomonoff's ideas in a concept
of a maximal semimeasure semicomputable from below (see also
Li and Vit{\'a}nyi~\cite{LiV97},
Section 4.5, Shen et al.~\cite{She2007}).
Levin proved that there exists a maximal to within a multiplicative
positive constant factor semimeasure $M$ semicomputable from below, i.e.
such that for every semimeasure $P$ semicomputable from below
a positive constant $c$ exists such that the inequality
\begin{equation}\label{M-ineq}
cM(x)\ge P(x)
\end{equation}
holds for all $x$. The semimeasure $M$ is called {\it the a priory}
or universal semimeasure.

For any semimeasure $Q$, its {\it support set} $E_Q$ is a set
of all infinite sequences $\omega$ such that $Q(\omega^n)>0$
for all $n$, i.e., $E_Q=\cup_{Q(x)>0}\Gamma_x$.

A function $P$ is a measure if (\ref{semi-1}) holds, where
both inequality signs $\le$ are replaced on $=$. Any function $P$ satisfying
(\ref{semi-1}) (with equalities) can be extended on all Borel subsets
of $\Omega$ if we define $P(\Gamma_x)=P(x)$ in $\Omega$,
where $x\in\Xi$ and $\Gamma_x=\{\omega\in\Omega:x\subseteq\omega\}$;
after that, we use the standard method for extending $P$ to all Borel
subsets of $\Omega$. By simple set in $\Omega$ we mean a union of
intervals $\Gamma_x$ from a finite set.

A measure $P$ is computable if it is, at one time, lower and upper
semicomputable.

For technical reasons, for any semimeasure $P$, we consider the maximal
measure $\bar P$ such that $\bar P\le P$. This measure satisfies
\begin{eqnarray*}
\bar P(x)=\inf_n\sum\limits_{l(y)=n,x\subseteq y}P(y).
\end{eqnarray*}
In general, the measure $\bar P$ is noncomputable (and it is not a
probability measure). By (\ref{M-ineq}), for each lower
semicomputable semimeasure $P$, the inequality $c\bar M(A)\ge\bar
P(A)$ holds for every Borel set $A$, where $c$ is a positive
constant.

In the manner of Levin's papers~\cite{Lev74,LevV78,Lev84,ZvL70}
(see also~\cite{Vyu82}), we consider combinations of
probabilistic and deterministic processes as the most general
class of processes for generating data. With any probabilistic
process some computable probability distribution can be
assigned. Any deterministic process is realized by means of an
algorithm. Algorithmic processes transform sequences generated
by probabilistic processes into new sequences. More precise, a
probabilistic computer is a pair $(P,F)$, where $P$ is a
computable probability distribution, and $F$ is a Turing
machine supplied with an additional input tape. In the process
of computation this machine reads on this tape a sequence
$\omega$ distributed according to $P$ and produces a sequence
$\omega'=F(\omega)$ (A correct definition see
in~\cite{Lev74, LiV97, She2007,Vyu82}). So,
we can compute the probability
\begin{eqnarray*}
Q(x)=P\{\omega\in\Omega: x\subseteq F(\omega)\}
\label{rep=1}
\end{eqnarray*}
that the result $F(\omega)$ of the computation begins with a
finite sequence $x$.
It is easy to see that $Q(x)$ is a semimeasure semicomputable from below.

Generally, the semimeasure $Q$ can be not
a probability distribution on $\Omega$,
since $F(\omega)$ may be finite for some infinite $\omega$.

The converse result is proved in Zvonkin and Levin~\cite{ZvL70}: for every
semimeasure $Q(x)$ semicomputable from below a probabilistic computer
$(L,F)$ exists such that
$$
Q(x)=L\{\omega\vert x\subseteq F(\omega)\},
$$
for all $x$, where $L(x)=2^{-l(x)}$ is the uniform probability
distribution on the set of all binary sequences.

Analogously, for any Borel set $A\subseteq\Omega$ consisting of infinite
sequences, we consider the probability
\begin{eqnarray}
Q(A)=L\{\alpha\in\Omega:F(\alpha)\in A\}
\label{M-ineq1}
\end{eqnarray}
of generating a sequence $\omega\in A$ by means of a probabilistic
computer $F$. Obviously, we have $c\bar M(A)\ge Q(A)$ for all such $A$,
where $c$ is a positive constant.

Therefore, by (\ref{M-ineq}) and (\ref{M-ineq1}) $M(x)$ and $M(A)$ define
universal upper bounds of the probability of generating $x$ and
$\omega\in A$ by probabilistic computers.

We distinguish between subsets of $\Omega$ of $\bar M$-measure $0$
and subsets of positive measure $\bar M$. If $\bar M(A)=0$ then the probability
of generating a sequence $\omega\in A$ by means of any probabilistic computer
is equal to $0$. 

The simplest example of a set of $\bar M$-measure $0$
is $A=\{\omega\}$, where $\omega$ is a non-computable sequence. Indeed,
if $\bar M\{\omega\}>0$ then there exist a rational $r>0$ such that
$M(\omega^n)>r$ for all $n$. Obviously, there are only finite number
of uncomparable strings $x$ such that $M(x)>r$. Then there exists an $k$
such that $\omega^k\subseteq x$ and $M(x)>r$ imply $x\subseteq\omega$.
We can compute each bit of $\omega$ by enumerating all such $x$.

The sets of $\bar M$-measure $0$ were described by Levin~\cite{Lev74, Lev84}
in terms of quantity of information.

We refer readers to Li and Vit{\'a}nyi~\cite{LiV97} and to
Shen et al.~\cite{She2007} for the theory of algorithmic randomness.
We use definition of a random sequence in terms of
universal probability. Let $P$ be some computable measure in $\Omega$.
The deficiency of randomness of a sequence $\omega\in\Omega$
with respect to $P$ is defined as
\begin{eqnarray}\label{def-1}
d(\omega|P)=\sup\limits_n\frac{M(\omega^n)}{P(\omega^n)},
\end{eqnarray}
where $\omega^n=\omega_1\omega_2\dots\omega_n$.
This definition leads to the same class of random sequences as the original
Martin-L\"of~\cite{Mar66} definition.
Let $R_P$ be the set of all infinite binary sequences random with respect to
a measure $P$
$$
R_P=\{\omega\in\Omega:d(\omega|P)<\infty\}.
$$
We also consider {\it parametric families} of probability distributions
$P_\theta(x)$, where $\theta$ is a real number;
we suppose that $\theta\in [0,1]$. An example of such family is
the Bernoulli family $B_\theta(x)=\theta^{k}(1-\theta)^{n-k}$,
where $n$ is the length of $x$ and $k$ is the number of ones in it.

We associate with a binary sequence $\theta_1\theta_2\dots$ a real number
with the binary expansion
$0.\theta_1\theta_2\dots$. When the sequence $\theta_1\theta_2\dots$ is
computable or random with respect to some measure we say that the number
$0.\theta_1\theta_2\dots$ is computable or random with respect to
the corresponding measure in $[0,1]$.

We consider probability distributions $P_\theta$ computable with respect to
a parameter $\theta$. Informally, this means that there exists an algorithm
enumerating all triples $(x,r_1,r_2)$, where $x\in\Xi$ and 
$r_1,r_2$ are rational numbers, such that $r_1<P_\theta(x)<r_2$. 
This algorithm uses an infinite sequence $\theta$
as an additional input; if some triple $(x,r_1,r_2)$ is enumerated by this
algorithm then only a finite initial fragment of $\theta$ was used in the
process of computation (for correct definition, see also
Shen et al.~\cite{She2007} and Vovk and V'yugin~\cite{VoV93}).

Analogously, we consider parametric lower semicomputable semimeasures.
It can be proved that there exist a universal parametric
lower semicomputable semimeasure $M_\theta$. This means that for each
parametric lower semicomputable semimeasure $R_\theta$ there exists
a positive constant $C$ such that $CM_\theta(x)\ge R_\theta(x)$ for all $x$
and $\theta$.

The corresponding definition of randomness with respect to a family
$P_\theta$ is
obtained by relativization of (\ref{def-1}) with respect to $\theta$
\begin{eqnarray*}\label{def-1a}
d_\theta(\omega)=\sup\limits_n\frac{M_\theta(\omega^n)}{P_\theta(\omega^n)}
\end{eqnarray*}
(see also~\cite{Lev73}). This definition leads to the same class of
random sequences as the original Martin-L\"of~\cite{Mar66} definition
relitivized with respect to a parameter $\theta$.

For any $\theta$, let
$$
I_\theta=\{\omega\in\Omega:d_\theta(\omega)<\infty\}
$$
be the set of all infinite binary sequences random with respect to
the measure $P_\theta$. In case of Bernoulli family, we call elements
of this set $\theta$-{\it Bernoulli sequences}.

\section{Randomness with respect to a parameter family}

We need some statistical notions (see Cox and Hinkley~\cite{CoH74}).
Let $P_\theta$ be some computable parametric family of probability
distributions. A function $\hat\theta(x)$ from $\Xi$ to $[0,1]$ is called
{\it an estimator}. An estimator $\hat\theta$ is called 
{\it strictly consistent}
if for each parameter value $\theta$ for $P_\theta$-almost all $\omega$,
$$
\hat\theta(\omega^n)\to\theta
$$
as $n\to\infty$.

Let $\epsilon$ and $\delta$ be rational numbers.
An estimator $\hat\theta$ is called {\it effectively strictly consistent}
if there exists a computable function $N(\epsilon,\delta)$ such that
for each $\theta$ for all $\epsilon$ and $\delta$
\begin{eqnarray}\label{con-alm}
P_\theta\{\omega\in\Omega:\sup\limits_{n\ge N(\epsilon,\delta)}
|\hat\theta(\omega^n)-\theta|>\epsilon\}\le\delta
\end{eqnarray}
The strong law of large numbers Borovkov~\cite{Bor74} (Chapter 5)
$$
B_\theta\left\{\sup\limits_{k\ge n}\left|\frac{1}{k}
\sum\limits_{i=1}^k\omega_i-\theta\right|
\ge\epsilon\right\}<\frac{1}{\epsilon^4 n}
$$
shows that the function
$\hat\theta(\omega^n)=\frac{1}{n}\sum\limits_{i=1}^n\omega_i$ is
a computable strictly consistent estimator for the Bernoulli family $B_\theta$.

\begin{proposition}\label{cons-1}
For any effectively strictly consistent estimator $\hat\theta$,
$$
\lim\limits_{n\to\infty}\hat\theta(\omega^n)=\theta
$$
for each $\omega\in I_\theta$.
\end{proposition}
{\it Proof}. Assume an infinite sequence $\omega$
be Martin-L\"of random with respect to $P_\theta$ for some $\theta$.

At first, we prove that
$\lim\limits_{n\to\infty}\hat\theta(\omega^n)$ exists.
Let for $j=1,2,\dots$,
$$
W_j=\{\alpha\in\Omega:(\exists n,k\ge N(1/j,2^{-(j+1)}))
|\hat\theta(\alpha^n)-\hat\theta(\alpha^k)|>1/j\}.
$$
By (\ref{con-alm}) for any $\theta$, $P_\theta(W_j)<2^{-j}$ for all $j$.
Define $V_i=\cup_{j>i}W_j$ for all $i$.
By definition for any $\theta$, $P_\theta(V_i)<2^{-i}$ for all $i$.
Also, any set $V_i$ can be represented as a recursively enumerable
union of intervals of type $\Gamma_x$. To reduce this definition of
Martin-L\"of test to the definition of the test (\ref{def-1}) define
a sequence of uniform lower semicomputable parametric semimeasures
\[
R_{\theta,i}(x)=
  \left\{
    \begin{array}{l}
      2^i P_\theta(x) \mbox{ if } \Gamma_x\subseteq V_i
    \\
      0 \mbox{ otherwise }
    \end{array}
  \right.
\]
and consider the mixture
$R_\theta(x)=\sum\limits_{i=1}^\infty \frac{1}{i(i+1)}R_{\theta,i}(x)$.

Suppose that $\lim\limits_{n\to\infty}\hat\theta(\omega^n)$
does not exist. Then for each sufficiently large $j$,
$|\hat\theta(\omega^n)-\hat\theta(\omega^k)|>1/j$ for
infinitely many $n$ and $k$. This implies that $\omega\in V_i$ for all $i$,
and then for some positive constant $c$,
$$
d_\theta(\omega)=\sup\limits_n\frac{M_\theta(\omega^n)}{P_\theta(\omega^n)}\ge
\sup\limits_n\frac{R_\theta(\omega^n)}{cP_\theta(\omega^n)}=\infty,
$$
i.e., $\omega$ is not Martin-L\"of random with respect to $P_\theta$.

Suppose that $\lim\limits_{n\to\infty}\hat\theta(\omega^n)\not =\theta$.
Then the rational numbers $r_1$, $r_2$ exist such that
$r_1<\lim\limits_{n\to\infty}\hat\theta(\omega^n)<r_2$ and
$\theta\not\in [r_1,r_2]$. Since the estimator $\hat\theta$ is consistent,
$P_\theta\{\alpha:r_1<\lim\limits_{n\to\infty}\hat\theta(\alpha^n)<r_2\}=0$,
and we can effectively (using $\theta$) enumerate an infinite sequence
of positive integer numbers $n_1<n_2<\dots$ such that for
$$
W'_j=\cup\{\Gamma_x:l(x)\ge n_j, r_1<\hat\theta(x)<r_2\},
$$
we have $P_\theta(W'_j)<2^{-j}$ for all $j$.
Define $V'_i=\cup_{j>i}W'_j$ for all $i$. We have $P_\theta(V'_i)\le 2^{-i}$
and $\omega\in V'_i$ for all $i$.
Then $\omega$ can not be Martin-L\"of random with respect to $P_\theta$.
These two contradictions obtained above prove the proposition.
$\triangle$

The following theorem generalizes the simplest example of a set of
$\bar M$-measure $0$ presented in Section~\ref{prel-1}. It
can be interpreted such that $P_\theta$-random sequences
with ``a prespecified'' noncomputable parameter $\theta$ can not be
obtained in any combinations of stochastic and deterministic processes.

\begin{theorem}\label{th-1} Assume a computable parametric family $P_\theta$
of probability distributions has an effectively strictly consistent
estimator. Then for each $\theta$,
$\bar M(I_\theta)>0$ if and only if $\theta$ is computable.
\end{theorem}
{\it Proof}. If $\theta$ is computable then the probability distribution
$P_\theta$ is also computable and by (\ref{M-ineq})
$c\bar M(I_\theta)\ge P_\theta(I_\theta)=1$, where $c$ is a positive constant.

The proof of the converse assertion is more complicated.
Assume $\bar M(I_\theta)>0$. There exists a simple set $V$ (a union of
a finite set of intervals) and a rational number $r$ such that
$\frac{1}{2}\bar M(V)<r<\bar M(I_\theta\cup V)$. For any finite set
$X\subseteq\Xi$, let $\bar X=\cup_{x\in X}\Gamma_x$.

Let $n$ be a positive integer number. When we compute a rational
approximation $\theta_n$ of $\theta$ up to $\frac{1}{2n}$ as follows.
Using the exhaustive search, we find a finite set $X_n$ of pairwise
incomparable finite sequences of length $\ge N(1/n, 2^{-n})$ such that
\begin{eqnarray}
\bar X_n\subseteq V,\mbox{ }\sum\limits_{x\in X_n}M(x)>r,
\nonumber
\\
|\hat\theta(x)-\hat\theta(x')|\le\frac{1}{2n}
\label{diff-1}
\end{eqnarray}
for all $x,x'\in X_n$.
If any such set $X_n$ will be found, we put $\theta_n=\hat\theta(x)$,
where $x$ is the minimal element of $X_n$
with respect to some natural (lexicographic)
ordering of all finite binary sequences.

Let us prove that for each $n$ some such set $X_n$ exists. Since
$\bar M(I_\theta\cap V)>r$, there exists a closed (in the topology defined
by intervals $\Gamma_x$) set $E\subseteq I_\theta\cap V$ such that
$\bar M(E)>r$. Consider the function
$$
f_k(\omega)=\inf\{n:n\ge k,|\hat\theta(\omega^n)-\theta|\le\frac{1}{4n}\}.
$$
By Proposition~\ref{cons-1} this function is continuous on $\Omega$ and,
since the set $E$ is compact, it is bounded on $E$. Hence, for each $k$,
there exists a finite set $X\subseteq\Xi$ consisting of pairwise incomparable
sequences of length $\ge k$ such that $E\subseteq\bar X$ and
$|\hat\theta(x)-\hat\theta(x')|\le\frac{1}{2n}$ for all $x,x'\in X$.
Since $E\subseteq\bar X$, we have $\sum\limits_{x\in X}M(x)>r$.
Therefore, the set $X_n$ can be found by exhaustive search.

\begin{lemma}\label{lem-1} For any Borel set $V\subseteq\Omega$,
$\bar M(V)>0$ and $V\subseteq I_\theta$ imply $P_\theta(V)>0$.
\end{lemma}
{\it Proof}. By definition of $M_\theta$ each computable parametric measure
$P_\theta$ is absolutely continuous with respect to the measure
$\bar M_\theta$, and so, we have representation
\begin{equation}\label{Rad-1}
P_\theta(X)=\int\limits_{X}\frac{dP_\theta}{d\bar M_\theta}(\omega)
d\bar M_\theta(\omega),
\end{equation}
where $\frac{dP_\theta}{d\bar M_\theta}(\omega)$ is the
Radon-Nicodim derivative; it exists for $\bar M_\theta$-almost all $\omega$.

By definition we have for $\bar M_\theta$-almost all $\omega\in I_\theta$
\begin{equation}\label{Rad-2}
\frac{dP_\theta}{d\bar M_\theta}(\omega)=\lim\limits_{n\to\infty}
\frac{P_\theta}{\bar M_\theta}(\omega^n)\ge\liminf\limits_{n\to\infty}
\frac{P_\theta}{\bar M_\theta}(\omega^n)\ge C_{\theta,\omega}>0.
\end{equation}

By definition $c_\theta\bar M_\theta(X)\ge\bar M(X)$ for all Borel sets $X$,
where $c_\theta$ is some positive constant (depending on $\theta$).
Then by (\ref{Rad-1}) and (\ref{Rad-2}) the inequality $\bar M(X)>0$
implies $P_\theta(X)>0$ for each Borel set $X$.
$\triangle$

We rewrite (\ref{con-alm}) in the form
\begin{eqnarray}\label{set-1}
E_n=\{\omega\in\Omega:\sup\limits_{N\ge N(1/(2n),2^{-n})}
|\hat\theta(\omega^N)-\theta|\ge\frac{1}{2n}\}.
\end{eqnarray}
By definition $P_\theta(E_n)\le 2^{-n}$ for all $n$. We prove that
$X_n\not\subseteq E_n$ for almost all $n$. Suppose that the opposite
assertion holds. Then there exists an increasing infinite
sequence of positive integer numbers $n_1,n_2\dots$ such that
$X_{n_i}\subseteq E_{n_i}$ for all $i=1,2,\dots$. This implies
$P_\theta(X_{n_i})\le 2^{-n_i}$ for all $i$. For any $k$,
define $U_k=\cup_{i\ge k}X_{n_i}$. Clearly, we have for all $k$,
$\bar M(\bar U_k)>r$ and
$P_\theta(\bar U_k)\le\sum\limits_{i\ge k}2^{-n_i}\le 2^{-n_k+1}$.
Let $U=\cap U_k$. Then $P_\theta(U)=0$ and
$\bar M(U)\ge r>\frac{1}{2}\bar M(V)$.
From $U\subseteq V$ and $\bar M(I_\theta\cap V)>\frac{1}{2}\bar M(V)$
the inequality $\bar M(I_\theta\cap U)>0$ follows. Then the set
$I_\theta\cap U$ consists of $P_\theta$-random sequences,
$P_\theta(I_\theta\cap U)=0$ and $\bar M(I_\theta\cap U)>0$.
This is a contradiction with Lemma~\ref{lem-1}.

Assume $X_n\not\subseteq E_n$ for all $n\ge n_0$. Let also,
a finite sequence $x_n\in X_n$ is defined such that
$$
\Gamma_{x_n}\cap (\Omega\setminus E_n)\not =\emptyset.
$$
Then from $l(x_n)\ge N(\frac{1}{2n},2^{-n})$ the inequality
$$
\left|\frac{1}{l(x_n)}
\sum\limits_{i=1}^{l(x_n)} (x_n)_i-\theta\right|<\frac{1}{2n}
$$
follows. By (\ref{diff-1}) we obtain $|\theta_n-\theta|<\frac{1}{n}$.
This means that the real number $\theta$ is computable.
Theorem is proved. $\triangle$

Let $Q$ be a computable probability distribution on $\theta$s
(i.e., on the set $\Omega$). Then the Bayesian mixture with respect to
the prior $Q$
\begin{eqnarray*}
P(x)=\int P_\theta(x)dQ(\theta)
\end{eqnarray*}
is also computable probability distribution.

Recall that $R_Q$ is the set of all infinite sequences Martin-L\"of random
with respect to a computable probability measure $Q$.
Obviously, $P(\cup_{\theta\in R_Q}I_\theta)=1$, and then
$\bar M(\cup_{\theta\in R_Q}I_\theta))>0$.
Moreover, it follows from Corollary 4 of Vovk and V'yugin~\cite{VoV93}
\begin{theorem}
For any computable measure $Q$, a sequence $\omega$ is random with respect
to the Bayesian mixture $P$ if and only if $\omega$ is random with respect to
a measure $P_\theta$ for some $\theta$ random with respect to the measure $Q$;
in other words,
$$
R_P=\cup_{\theta\in R_Q}I_\theta.
$$
\end{theorem}
Notice that each computable $\theta$ is
Martin-L\"of random with respect to the computable probability distribution
concentrated on this sequence.

\section{Randomness with respect to non-random parameters}

We show in this section that the Bayesian
approach is insufficient to cover all possible ``meaningful''
cases: a probabilistic machine can be
constructed, which with probability close to one outputs a random
$\theta$-Bernoulli sequence, where the parameter $\theta$ is not
random with respect to each computable probability distribution.

Let ${\cal P}(\Omega)$ be the set of all computable probability 
measures on $\Omega$ and let
$$
{\cal S}=\cup_{P\in {\cal P}(\Omega)}R_P
$$
be the set of all sequences Martin-L\"of random with respect to 
computable probability measures. We call these sequences - {\it stochastic}.
Let ${\cal S}^c$ be a complement of $\cal S$ - the set of {\it non-stochastic}
sequences.

An infinite binary sequence $\alpha$ is Turing reducible to an
infinite binary sequence sequence $\beta$ if
$\alpha=F(\beta)$ for some computable operation $F$; we denote this 
$\alpha\le_T\beta$. Two infinite sequences $\alpha$ and $\beta$ are 
Turing equivalent if $\alpha\le_T\beta$ and $\beta\le_T\alpha$.
Let 
\begin{eqnarray}\label{cl-s}
Cl({\cal S})=\{\alpha:\exists\beta(\beta\in {\cal S}\&\beta\le_T\alpha)\}.
\end{eqnarray} 
The complement of the set (\ref{cl-s}),
$Cl({\cal S})^c=\Omega\setminus Cl({\cal S})$, consists of sequences
non-random with respect to all computable probability disributions, i.e.,
$Cl({\cal S})^c\subseteq {\cal S}^c$; moreover, it consists of sequences
which can not be Turing equivalent to stochastic sequences. Also, no stochastic
sequence can be Turing reducible to a sequence from $Cl({\cal S})^c$. 

V'yugin~\cite{Vyu76},~\cite{Vyu82} proved that $\bar M(Cl({\cal S})^c)>0$.

\begin{theorem}\label{pr-1}
For any $\epsilon$, $0<\epsilon<1$, a lower
semicomputable semimeasure $Q$ exists such that
$\bar Q(E_Q)>1-\epsilon$ and $E_Q\subseteq Cl({\cal S})^c$.
\end{theorem}
For completnees of presentation we give in Appendix~\ref{proof-1} 
a new simplified proof of this theorem.

We show that this result can be extended on parameters of the Bernoulli family.
\begin{theorem}\label{n-1} Let $I_\theta$ be the set of all
$\theta$-Bernoulli sequences. Then
$$
\bar M(\cup_{\theta\in Cl({\cal S})^c}I_\theta)>0.
$$
In terms of probabilistic computers, for any $\epsilon$, $0<\epsilon<1$,
a probabilistic machine $(L,F)$ can be constructed, which
with probability $\ge 1-\epsilon$ generates an $\theta$-Bernoulli sequence,
where $\theta\in Cl({\cal S})^c$ (i.e., $\theta$ is nonstochastic).
\end{theorem}
{\it Proof}. For any $\epsilon>0$, $0<\epsilon<1$,
we define a lower semicomputable semimeasure $P$ such that
$$
\bar P(\cup_{\theta\in Cl({\cal S})^c}I_\theta)>1-\epsilon.
$$
The proof of the theorem is based on Theorem~\ref{pr-1}. 

Let $Q$ be the semimeasure defined in this theorem.
For any $\omega\not\in E_Q$ we have $Q(\omega^n)=0$ for
all sufficiently large $n$. 
For the measure
\begin{eqnarray}\label{mes-1}
R^-(x)=\int B_\theta(x)d\bar Q(\theta),
\end{eqnarray}
where $B_\theta$ is the Bernoulli measure,
we have $R^-(\Omega)>1-\epsilon$ by Theorem~\ref{pr-1}, and\\
$R^-(\cup_{\theta\in Cl({\cal S})}I_\theta)=0$.

Unfortunately, we can not conclude
that $c\bar M\ge R^-$ for some constant $c$, since the measure $R^-$
is not represented in the form $R^-=\bar P$
for some lower semicomputable semimeasure $P$. To overcome this problem,
we consider some semicomputable approximation of this measure.

For any finite binary sequences $\alpha$ and $x$, let
$B^-_{\alpha}(x)=(\theta^-)^K(1-\theta^+)^{N-K}$, where $N$ is the
length of $x$ and $K$ is the number of ones in it, $\theta^-$ is
the left side of the subinterval corresponding to the sequence
$\alpha$ and $\theta^+$ is its right side. By definition
$B^-_{\alpha}(x)\le B_\theta(x)$ for all $\theta^-\le\theta\le\theta^+$.

Let $\epsilon$ be a rational number.
Let $Q^s(x)$ be equal to the maximal rational number $r<Q(x)$ computed in $s$
steps of enumeration of $Q(x)$ from below. Using Theorem~\ref{pr-1}, we can
define for $n=1,2,\dots$ and for each $x$ of length $n$ a computable sequence
of positive integer numbers $s_x\ge n$ and a sequence of finite
binary sequences $\alpha_{x,1},\alpha_{x,2},\dots\alpha_{x,k_x}$
of length $\ge n$ such that the function $P(x)$ defined by
\begin{eqnarray}\label{mes-3}
P(x)=\sum\limits_{i=1}^{k_x}B^-_{\alpha_{x,i}}(x)Q^{s_x}(\alpha_{x,i})
\end{eqnarray}
is a semimeasure, i.e., such that condition (\ref{semi-1}) holds for all $x$,
and such that
\begin{eqnarray}\label{no-3}
\sum\limits_{l(x)=n}P(x)>1-\epsilon
\end{eqnarray}
holds for all $n$. These sequences exist, since the limit function $R^-$
defined by (\ref{mes-1}) is a measure satisfying $R^-(\Omega)>1-\epsilon$.

By definition the semimeasure $P(x)$ is lower semicomputable. Then
$cM(x)\ge P(x)$ holds for all $x\in\Xi$, where $c$ is a positive
constant.

To prove that $\bar P(\Omega\setminus\cup_{\theta}I_\theta)=0$
we consider some probability  measure $Q^+\ge Q$.
Since (\ref{semi-1}) holds, it is possible
to define some noncomputable measure $Q^+$ satisfying these properties
in many different ways.
Define the mixture of the Bernoulli measures with respect to $Q^+$
\begin{eqnarray}\label{mes-2}
R^+(x)=\int B_\theta(x)dQ^+(\theta).
\end{eqnarray}
By definition $R^+(\Omega\setminus\cup_{\theta}I_\theta)=0$.
Using definitions (\ref{mes-3}) and (\ref{mes-2}),
it can be easily proved that $\bar P\le R^+$. Then
$\bar P(\Omega\setminus\cup_{\theta}I_\theta)=0$. By Theorem~\ref{pr-1}
$Cl({\cal S})\subseteq\Omega\setminus E_Q$, and then $\bar Q(Cl({\cal S}))=0$.
By (\ref{mes-3}) we have $\bar P(\cup_{\theta\in Cl({\cal S})}I_\theta)=0$. 
By (\ref{no-3}) we have $\bar P(\Omega)>1-\epsilon$. Then
$\bar P(\cup_{\theta\in Cl({\cal S})^c}I_\theta)>1-\epsilon$.
Therefore, $\bar M(\cup_{\theta\in Cl({\cal S})^c}I_\theta)>0$.
$\triangle$

\appendix

\section{Proof of Theorem~\ref{pr-1}}\label{proof-1}

Recall that $E_Q$ is the support set of a semimeasure $Q$.
In that follows we define a semicomputable semimeasure $Q$ such
that 
\begin{itemize}
\item{1)}
$\bar Q(E_Q)>0$;
\item{2)}
for each $\omega\in E_Q$ and for each computable operation $F$ 
such that $F(\omega)$ is infinite, the
sequence $F(\omega)$ is not Martin-L\"of random with respect to
the uniform probability measure $L$ on $\Omega$.
\end{itemize} 

By Theorem 4.2 from \cite{ZvL70} for each computable measure $P$ on $\Omega$,
there exist two computable operations $F$ and $G$ such that
\begin{itemize}
\item{3)} $F(\omega)\in\Omega$ for each $\omega$ random with
    respect to $L$, and $G(F(\omega))=\omega$;
\item{4)} for each sequence $\omega$ random with respect to $P$
    (and such that $P\{\omega\}=0$), the sequence
    $G(\omega)$ is random with respect to $L$.
\end{itemize}
By 1)-4) each sequence $\omega\in E_{Q}$ can not be Martin-L\"of random
with respect to any computable probability measure $P$.

We will construct a semicomputable semimeasure $Q$ as a some
sort of network flow. We define an infinite network on the base
of the infinite binary tree. This networt has no sink; 
the top of the tree (empty sequence) is the source.

Each $x\in\Xi$ defines two edges
$(x,x0)$ and $(x,x1)$ of length one. In the construction below
we will add to the network extra edges $(x,y)$ of length
$>1$, where $x,y\in\Xi$, $x\subseteq y$ and $y\not =x0, x1$.
By the length of the edge $(x,y)$ we mean the number
$l(y)-l(x)$. For any edge $\sigma=(x,y)$ we denote by
$st(\sigma)=x$ its starting vertex and by $ter(\sigma)=y$ its
terminal vertex. A computable function $q(\sigma)$ defined on
all edges of length one and on all extra edges and taking
rational values is called {\it a network} if for all $x\in\Xi$
\begin{eqnarray*} \label{net-1}
\sum\limits_{\sigma:\mbox{ }st(\sigma)=x} q(\sigma)\le 1.
\end{eqnarray*}
Let $G$ be the set of all extra edges of the network $q$
(it is a part of the domain of $q$).
By $q$-{\it flow } we mean the minimal semimeasure $P$ such that
$P\ge R$, where the function $R$ is defined by the following
recursive equations
\begin{eqnarray}
R(\lambda)=1;
\nonumber
\\
R(y)=\sum\limits_{\sigma:\mbox{ }ter(\sigma)=y}q(\sigma)R(st(\sigma))
\label{net-base-2}
\end{eqnarray}
for $y\not =\lambda$. It is easy to see that this semimeasure $P$
is lower semicomputable if $q$ is computable.

A network $q$ is called {\it elementary} if the set of extra edges
is finite and $q(\sigma)=1/2$ for almost all edges of unit length.
For any network $q$, we define the {\it network flow delay} function
($q$-delay function)
\begin{eqnarray*}
d(x)=1-q(x,x0)-q(x,x1).
\end{eqnarray*}
The construction below works with all programs $i$ computing
the operations $F_i(x)$.
\footnote {The existence of the
effectively computable sequence $\{F_{i}\}$ such that for each
computable operation $F$, $F=F_{i}$ for some $i$ is proved
in~\cite{Rog67}.}
We define
some function $p(n)$ such that for each positive integer number
$m$ we have $p(n)=m$ for infinitely many $n$. For example, we
can define $p(\langle m,k\rangle)=m$ and $p'(\langle
m,k\rangle)=k$ for all $m$ and $k$, where $\langle m,k\rangle$
is some computable one-to-one enumeration of all pairs of
nonnegative integer numbers. Then for each step $n$ we compute
$\langle i,s\rangle=p(n)$, where $i$ is a program and $s$ is a
number (we call $s$ number of a session); so, $i=p(p(n))$ and $s=p'(p(n))$.

Let a program $i$, a number $s$, finite binary sequences $x$
and $y$, an elementary network $q$, and a nonnegative integer
number $n$ be given. Define $B(\langle i,s\rangle,x,y,q,n)$ be
{\it true} if the following conditions hold
\begin{itemize}
\item{(i)}
$l(y)=n$, $x\subseteq y$,
\item{(ii)}
$d(y^k)<1$ for all $k$, $1\le k\le n$, where $d$ is the $q$-delay function
and $y^k=y_1\dots y_k$;
\item{(iii)} $l(F_i(y))>\langle x,s\rangle$.
\end{itemize}
Let $B(\langle i,s\rangle,x,y,q,n)$ be {\it false}, otherwise. Define
\begin{eqnarray*}
\beta(x,q,n)=\min\{y:p(l(y))=
p(l(x)),B(\langle p(p(l(x))),p'(p(l(x))\rangle,x,y,q,n)\}
\end{eqnarray*}
Here $p(p(l(x))$ is a program and $p'(p(l(x))$ is a number of session;
$\min$ is considered for lexicographical ordering of strings; we suppose
that $\min\emptyset$ is undefined.
\begin{lemma}\label{num-hard}
For each computable operation $F_i$ and for each finite sequence
$x$ such that $F(\omega)\in\Omega$ for some infinite extension
$\omega$ of $x$ (i.e., $x\subseteq\omega$), $\beta(x,q,n)$ is
defined for all sufficiently large $n$ such that $p(p(n))=i$.
\end{lemma}
{\it Proof}. The needed sequence $y$ exists for all
sufficiently large $n$, since $l(F_{i}(\omega^{n}))>\langle
x,s\rangle$ holds for all sufficiently large $n$, $p(n)=\langle
i,s\rangle$.
$\triangle$

The goal of the construction below is the following. Each extra
edge $\sigma$ will be assigned to some task number $I=\langle
i,s\rangle$ such that $p(l(st(\sigma)))=p(l(ter(\sigma)))=I$. The
goal of the task $I$ is to define a finite set of extra edges
$\sigma$ such that for each infinite binary sequence $\omega$
one of the following conditions hold: either $\omega$ contains
some extra edge as a subword, or the network flow delay
function $d$ equals $1$ on some initial fragment of $\omega$.
For each extra edge $\sigma$ added to the network $q$,
$B(I,st(\sigma),ter(\sigma),q^{n-1},n)$ is true; it is false,
otherwise. Lemma~\ref{nontriv-1a} shows that $\bar
Q(E_Q)>1-\epsilon$, where $Q$ is the $q$-flow and $E_Q$ is
its support set.

{\bf Construction.}
Let $\rho(n)=(n+n_0)^2$ for some sufficiently large $n_0$ (the value $n_0$
will be specified below in the proof of Lemma~\ref{nontriv-1a}).

Using the mathematical induction by $n$, we define a sequence $q^n$
of elementary networks. Put $q^0(\sigma)=1/2$ for all edges $\sigma$
of length one.

Assume $n>0$ and a network $q^{n-1}$ is defined.
Let $d^{n-1}$ be the $q^{n-1}$-delay function
and let $G^{n-1}$ be the set of all extra edges. We suppose also that
$l(ter(\sigma))<n$ for all $\sigma\in G^{n-1}$.

Let us define a network $q^n$. At first, we define
a network flow delay function $d^n$ and a set $G^n$.

Let $w(I,q^{n-1})$ be equal to the minimal
$m$ such that $p(m)=I$ and $m>l(ter(\sigma))$ for each extra edge
$\sigma\in G^{n-1}$ such that $p(l(st(\sigma)))<I$.

The inequality $w(I,q^m)\not =w(I,q^{m-1})$
can be induced by some task $J<I$ that adds an extra edge $\sigma=(x,y)$
such that $l(y)>w(i,q^{m-1})$ and $p(l(x))=p(l(y))=J$.
Lemma~\ref{gen-tech-1} (below) will show that this can happen only at finitely
many steps of the construction.

The construction can be split up into three cases.

{\it Case 1}. $w(p(n),q^{n-1})=n$ (the goal of this part is
to start a new task $I=p(n)$ or to restart the existing task $I=p(n)$ if it
was destroyed by some task $J<I$ at some preceding step).

Put $d^n(y)=1/\rho(n)$ for $l(y)=n$ and define
$d^n(y)=d^{n-1}(y)$ for all other $y$. Put also $G^n=G^{n-1}$.

{\it Case 2.} $w(p(n),q^{n-1})<n$ (the goal of this part is to process
the task $I=p(n)$).

Let $C_n$ be the set of all $x$ such that $w(I,q^{n-1})\le l(x)<n$,
$0<d^{n-1}(x)<1$, the function $\beta(x,q^{n-1},n)$ is defined
\footnote
{
In particular, $p(l(x))=I$ and $l(\beta(x,q^{n-1},n))=n$.
}
and there is no extra edge $\sigma\in G^{n-1}$ such that $st(\sigma)=x$.

In this case for each $x\in C_n$ define $d^n(\beta(x,q^{n-1},n))=0$,
and for all other $y$ of length $n$ such that $x\subseteq y$ define
$$
d^n(y)=d^{n-1}(x)/(1-d^{n-1}(x)).
$$
Define $d^n(y)=d^{n-1}(y)$ for all other $y$.
We add an extra edge to $G^{n-1}$, namely, define
\begin{eqnarray*}
G^n=G^{n-1}\cup\{(x,\beta(x,q^{n-1},n)):x\in C_n\}.
\end{eqnarray*}

We say that the task $I=p(n)$ {\it adds} the extra edge
$(x,\beta(x,q^{n-1},n))$ to the network
and that all existing tasks $J>I$ are destroyed by the task $I$.

After Case 1 and Case 2, define for each edge $\sigma$ of unit length
$$
q^n(\sigma)=\frac{1}{2}(1-d^n(st(\sigma)))
$$
and $q^n(\sigma)=d^n(st(\sigma))$ for each extra edge $\sigma\in G^n$.

{\it Case 3}. Cases 1 and 2 do not hold.

Define $d^n=d^{n-1}$, $q^n=q^{n-1}$, $G^n=G^{n-1}$.

Using this construction, we define the network
$q=\lim\limits_{n\to\infty}q^n$, the network flow delay function
$d=\lim\limits_{n\to\infty}d^n$, and the set of extra edges $G=\cup_n G^n$.

The functions $q$ and $d$ are computable and the set $G$ is recursive
by their definitions. Let $Q$ denote the $q$-flow.

The following lemma shows that any task can add new extra edges only
at finite number of steps.

Let $G(I)$ be the set of all extra edges added by the task $I$,
$w(I,q)=\lim_{n\to\infty} w(I,q^n)$.
\begin{lemma} \label{gen-tech-1}
The set $G(I)$ is finite and $w(I,q)<\infty$ for all $I$.
\end{lemma}
{\it Proof.} Note that if $G(J)$ is finite for all $J<I$
then $w(I,q)<\infty$. Then we must prove that the set
$G(I)$ is finite for all $I$. Suppose that the opposite assertion holds.
Let $I$ be the minimal number such that $G(I)$ is infinite.
By choice of $I$ the sets $G(J)$ for all $J<I$ are finite.
Then $w(I,q)<\infty$.

By definition if $d(\omega^m)\not=0$ then $p_m=1/d(\omega^m)$
is a positive integer number. Besides, if $(\omega^n,y),
(\omega^m,y')\in G(I)$, where $n<m$ and $l(y)=m$, then $p_n>p_m$.
Hence, for each $\omega\in\Omega$ a maximal $m$ exists such that
$(\omega^m,y)\in G(I)$ for some $y$ or no such extra edge exists.
In the latter case put $m=w(I,q)$. Define $u(\omega)=1/d(\omega^m)$.

By the construction the integer valued function  $u(\omega)$ is constant
on the interval $\Gamma_{\omega^m}$. Hence, it is continuous
in the topology generated by such intervals. Since $\Omega$ is compact in this
topology, $u(\omega)$ is bounded. Then for some $m'$,
$u(\omega)=u(\omega^{m'})$ for all $\omega$. By the
construction if any extra edge of $I$th type was added to
$G(I)$ at some step then $d(y)>d(x)$ holds for some new pair
$(x,y)$ such that $x\subseteq y$. This is a contradiction
if $G(I)$ is infinite. $\triangle$

An infinite sequence $\alpha\in\Omega$ is called an $I$-{\it extension}
of a finite sequence $x$ if $x\subseteq\alpha$ and $B(I,x,\alpha^n,n)$
is true for almost all $n$.

A sequence $\alpha\in\Omega$ is called $I$-{\it closed}\mbox{ }
if $d(\alpha^n)=1$ for some $n$ such that $p(n)=I$, where $d$ is the
$q$-delay function. Note that if $\sigma\in G(I)$ is some extra
edge then $B(I,st(\sigma),ter(\sigma),n)$
is true, where $n=l(ter(\sigma))$.

\begin{lemma} \label{exten-1}
Assume for each initial fragment $\omega^n$ of an infinite sequence
$\omega$ some $I$-extension exists. Then either the sequence
$\omega$ will be $I$-closed in the process of the construction or
$\omega$ contains an extra edge of $I$th type (i.e. such that
$ter(\sigma)\subseteq\omega$ for some $\sigma\in G(I)$).
\end{lemma}
{\it Proof.} Assume a sequence $\omega$ is not $I$-closed. By
Lemma~\ref{gen-tech-1} the maximal $m$ exists such that $p(m)=I$
and $d(\omega^m)>0$. Since the sequence $\omega^m$ has an
$I$-extension and $d(\omega^k)<1$ for all $k$, by Case 2 of the construction a
new extra edge $(\omega^m,y)$ of $I$th type must be added to the
binary tree. By the construction $d(y)=0$ and $d(z)\not =0$ for all
$z$ such that $\omega^m\subseteq z$, $l(z)=l(y)$, and $z\not=y$.
By the choice of $m$ we have $y\subseteq\omega$. $\triangle$

Obviously, $Q(y)=0$ if and only if $q(\sigma)=0$ for some edge
$\sigma$ of unit length located on $y$ (this edge satisfies
$ter(\sigma)\subseteq y$ and $d(st(\sigma))=1$). Then the relation
$Q(y)=0$ is recursive and $E_Q=\Omega\setminus\cup_{d(x)=1}\Gamma_x$.
\begin{lemma} \label{nontriv-1a}
It holds $\bar Q(E_Q)>1-\epsilon$.
\end{lemma}
{\it Proof.} We bound $\bar Q(\Omega)$ from
below. For any $n$, let $q^n$ be the network defined at step
$n$, $R^{n}$ be defined by (\ref{net-base-2}), and $d^{n}$ be
the corresponding $q^n$-delay function. If $w(p(n),q^{n-1})=n$
(i.e., Case 1 holds at step $n$) then
\begin{eqnarray}
\sum\limits_{l(u)=n} d^{n}(u)R^{n}(u)=(n+n_{0})^{-2}\sum R^{n}(u)\le (n+n_{0})^{-2}.
\label{RR-2ab}
\end{eqnarray}

Assume Case 2 holds at the step $n$ and $x\in C_{n}$ such that
$(x,y)\in G$ for some $y$, $l(y)=n$. Since by the construction $d^n(y)=0$,
\begin{eqnarray}
\sum\limits_{l(z)=n, x\subseteq z}
d^{n}(z)R^{n}(z)\le\frac{d^{n-1}(x)}{(1-d^{n-1}(x))}
\sum\limits_{l(z)=n, x\subseteq z, z\not =y} R^{n}(z).
\label{RR-2a}
\end{eqnarray}
We have
\begin{eqnarray}
\sum\limits_{l(z)=n, x\subseteq z, z\not =y}
R^{n-1}(z)\le (1-d^{n-1}(x))R^{n-1}(x).
\label{RR-2aa}
\end{eqnarray}
By the construction
$R^{n}(z)=R^{n-1}(z)$ for $z$ such that $l(z)=n$, $x\subseteq z$, $z\not =y$.
Then
\begin{eqnarray}
\sum\limits_{l(z)=n, x\subseteq z} d^{n}(z)R^{n}(z)\le d^{n-1}(x)R^{n-1}(x).
\label{RR-2aaa}
\end{eqnarray}
By definition $\sum (n+n_{0})^{-2}\le\epsilon$.
After that, using (\ref{RR-2ab}) and (\ref{RR-2aaa}) we can
prove by the mathematical induction on $n$ that
$$
\bar Q(\Omega)=\inf\limits_{n}\sum\limits_{l(u)=n} Q(u)\ge
\inf\limits_{n}\sum\limits_{l(u)=n} R(u)\ge 1-\epsilon.
$$
Lemma is proved. $\triangle$

\begin{lemma} \label{nontriv-1b} For any infinite sequence
$\omega\in E_Q$ and for any computable operation $F$ if the
sequence $F(\omega)$ is infinite then it is not Martin-L\"of
random with respect to the uniform probability distribution.
\end{lemma}
{\it Proof.} Assume that $\omega$ is an infinite sequence and $F$
is a computable operation such that $F(\omega)$ is infinite.
Then $F_{i}=F$ for some $i$. Define
$$
U_{s}=\cup\{\Gamma_{\beta(x,q^{n-1},n)}:x\in C_{n}, p(n)=\langle i,s\rangle\},
$$
where $C_{n}$ is the set from Case 2 of the construction. By definition
$$
L(U_{s})=\sum\limits_{x\in C_{n}} 2^{-\langle x,s\rangle}\le 2^{-cs}
$$
for some positive constant $c$,
and $F_{i}(\omega)\in\cap_{s}U_{s}$. Therefore, the sequence
$F(\omega)$ is not Martin-L\"of random. Lemma~\ref{nontriv-1b}
and Theorem~\ref{pr-1} are proved.
$\triangle$

\end{document}